\documentclass[preprint]{elsarticle}
\usepackage{lineno,hyperref}
\modulolinenumbers[5]
\usepackage{amsfonts}
\usepackage{multirow}
\usepackage{soul,color}
\usepackage{subcaption} 

\journal{Journal of \LaTeX\ Templates}

\begin{document}

\begin{frontmatter}

\title{Common Sense Knowledge Learning for Open Vocabulary Neural Reasoning: A First View into Chronic Disease Literature
}

\author{Ignacio Arroyo-Fernández\fnref{myfootnote}\corref{mycorrespondingauthor}}
\fntext[myfootnote]{Universidad Tecnológica de la Mixteca $|$ Huajuapan de León, Oaxaca, México $|$ 69000}
\cortext[mycorrespondingauthor]{Corresponding author}
\ead{iaf@gs.utm.mx}

\author{José Armando Sánchez-Rojas\fnref{myfootnote1}}

\author{Arturo Téllez-Velázquez\fnref{myfootnote1}}
\author{Flavio Juárez-Martínez\fnref{myfootnote1}}

\author{Raúl Cruz-Barbosa\fnref{myfootnote1}}

\author{Enrique Guzmán-Ramírez\fnref{myfootnote1}}
\author{Yalbi I. Balderas-Martínez\fnref{myfootnote2}}
\fntext[myfootnote2]{Instituto Nacional de Enfermedades Respiratorias ``Ismael Cosío Villegas'' $|$ Ciudad de México $|$ 14080}



\begin{abstract}
In this paper, we address reasoning tasks from open vocabulary Knowledge Bases (openKBs) using state-of-the-art Neural Language Models (NLMs) with applications in scientific literature. 
For this purpose, self-attention based NLMs are trained using a common sense KB as a source task. The NLMs are then tested on a target KB for open vocabulary reasoning tasks involving scientific knowledge related to the most prevalent chronic diseases (also known as non-communicable diseases, NCDs). Our results identified NLMs that performed consistently and with significance in knowledge inference for both source and target tasks. Furthermore, in our analysis by inspection we discussed the semantic regularities and reasoning capabilities learned by the models, while showing a first insight into the potential benefits of our approach to aid NCD research.
\end{abstract}

\begin{keyword}
\texttt{elsarticle.cls}\sep \LaTeX\sep Elsevier \sep template
\MSC[2010] 00-01\sep  99-00
\end{keyword}

\end{frontmatter}


\section{Introduction}

The scientific community in multiple areas is now proactive in generating knowledge aimed at addressing the needs to improve public health and quality of life of the world population. Evidently, this knowledge is generated with the main aim of being persistent and reusable \cite{harper2018agbiodata}. Ironically, its evolution and volume makes it even more difficult to access, share and interpret it. We believe that generalizing access to knowledge of scientific domains relevant to societal challenges using Natural Language Processing (NLP) and Deep Learning (DL) methods is a promising approach to advance the aforementioned problem.


Nowadays, Knowledge Bases (KBs) are important pieces of technology that organize knowledge in a way that scientists from different areas can navigate the structure and meaning of the documented empiricism. KBs are commonly applied in organizing knowledge generated for different levels of specificity in different areas of Health Care and Biomedicine \cite{hunter2017knowledge}. On the one hand, there is a well known set of NLP techniques, called Information Extraction (IE, \cite{sheikhalishahi2019natural,yu2018artificial}), that have been applied to automatically generate KBs and Knowledge Graphs (KGs). On the other hand, Neural Network-Based Language Models (NLMs) are becoming a very useful set of methods that can take these KBs as training data and use them to perform semantic reasoning tasks such as Common Sense Reasoning (CSR) and Knowledge Base Completion (KBC) \cite{davis2015commonsense,socher2013reasoning}. 

The goal of the trained NLM in the KBC task is to infer the missing parts of semantic structures, i.e. the NLM learns from training samples of the form \texttt{\{Subject, Predicate, Object\}} (i.e. SPO triples, e.g. \texttt{\{music, is a, form of communication\}}). This includes the training triples edited by the model. For instance, given an incomplete SPO triple, i.e. the subject and the predicate phrases, \texttt{\{music, is a, \_\_\}}, the goal of the trained model is to infer the missing object phrase: \texttt{form of communication}. open vocabulary KBs (openKBs, for short) are generalizations of KBs, whose entries (SPO triples) are built from natural language phrases, rather than from fixed inventories of entities and relations.

The mentioned models and reasoning tasks compose a well-known methodology for knowledge processing. Nonetheless, its effectiveness is currently limited by (\textit{i}) the method used to build the KB, and (\textit{ii}) the generalization ability of the NLM in reasoning. IE methods, in particular Relation Extraction (RE), are the building blocks of KBs and KGs. However, they are currently limited by their low recall \cite{xu-etal-2013-open}, which is important because it directly affects the expressiveness and contextual information (very controlled syntactic structure and vocabulary) provided by the resulting KB \cite{cetto-etal-2018-graphene-context,shin2015incremental,popescul2003statistical,koller2007introduction,getoor2011learning}. The low recall of simple RE methods is due to the fact that the subject and object of a triple are limited to a predefined set of named entities, which are very special cases of noun phrases, and the predicate to a small set of predefined attributional or defining verbs.

NLMs based on Recurrent Neural Networks (RNNs) have shown limitations in reasoning tasks \cite{hochreiter1997long,zhou2015end}, particularly encoder-decoder architectures \cite{cho2014learning,chaudhari2021attentive}. These models have been improved with contextual alignment (i.e. attention) between source and target sequences \cite{bahdanau2015neural,luong-etal-2015-effective}, but recent developments have shown even better improvements, i.e. the Transformer encoder-decoder architecture \cite{vaswani2017attention}, which is a self-attention-based Neural Language Model. From these developments, pre-trained Transformer-based NLMs are becoming ubiquitous in multiple IE and reasoning tasks \cite{kenton2019bert,yang2019xlnet,floridi2020gpt}, however the need to prune and tweak them to make more efficient and specialized applications is also becoming apparent \cite{han-etal-2021-robust,lee2020biobert,meng2021mixture,pfeiffer2020adapterhub}. Moreover, they only consider general purpose vocabulary, which makes difficult their transferability to open domains. Openness for new domains allows selecting an arbitrary specific target (application) domain.

In this work, we address the limited flexibility of the current reasoning methodology based on KBs. We apply the obtained improvements to a reasoning task in the context of the research literature on Chronic Diseases (also known as Non-Communicable Diseases, NCDs). For this, we use an open vocabulary chronic disease KB, that we call OpenNCDKB, built using a state-of-the-art OpenIE on abstracts of articles retrieved from PubMed. Consequently, we trained a Transformer-based NLM to solve a set of similar source reasoning tasks (KBC) that involve general purpose vocabulary: the ConceptNet knowledge Graph and an openKB we built from multiple sources of Open Information Extractions (OpenIE \cite{etzioni2008open,fader2011identifying}). Using the models obtained for the source tasks, we observed their reasoning performance and capabilities on our OpenNCDKB. 
Our results showed that a significant number of SPO structures used as ground truth were predicted by the NLM. Other predictions were incorrect from the point of view of loss function and accuracy. However, we note that, rather than being due to ground-truth errors, from the point of view of language meaning, the view of these metrics could be biased towards accuracy (which is not peculiar to natural language), although this was effective in learning. We subsequently confirmed this using an STS-based hypothesis test of the predicted and true object sentence embeddings in the source and target tasks \cite{arroyo2019unsupervised,wang2021distributed}. 
This helped us to analyze by inspection the semantic regularities of some predictions taken at random. The apparent robustness of these regularities showed the potential benefits of the new knowledge inferred by the model from our OpenNCDKB, which suggested that such knowledge (not in the training data) needs to be validated through further research, rather than being directly discarded as a set of senseless predictions.

The rest of this paper is organized as follows. In Section \ref{sec:02_methodology} we expose our methodology. In Section \ref{sec:03_theoretical_background} we show the theoretical background. In Section \ref{sec:05_data_and_experimental_setup} we describe the data and the experimental setup. In Section \ref{sec:06_results_discussion} we report our results, together with the corresponding discussion. In Section \ref{sec:07_related_work} we describe prior work related to ours, and in Section \ref{sec:08_conclusions} we present our conclusions.

\section{Methodology}\label{sec:02_methodology}
\begin{figure}
    \centering
    \includegraphics[clip, trim=0.5cm 7.3cm 0.5cm 9.6cm, scale=0.5
    ]{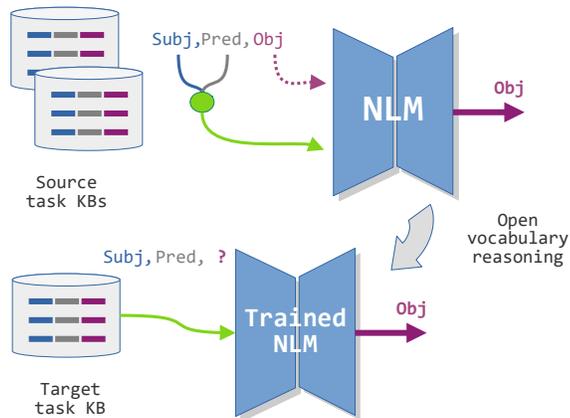}
    \caption{Our general methodology.}
    \label{fig:methodology}
\end{figure}

DL and NLP researchers recently tested Transformer-Based NLMs in general purpose CSR tasks \cite{bosselut2019comet,lin2019kagnet}. In sight of this progress, our research uses a methodological approach that can be especially useful for the semantic analysis of documents dealing with arbitrary but specialized topics, such as open domain scientific literature. Figure \ref{fig:methodology} shows the general methodology used in this work.

First, we trained Transformer-based encoder-decoder NLMs to solve combinations of two similar source reasoning tasks that involve general purpose domains: the ConceptNet KG in its Common Sense KB version, and an openKB we built from multiple sources using OpenIE. The target specialized domains were included in an open vocabulary Chronic Disease KB (OpenNCDKB) we built using a state-of-the-art OpenIE method that extracts open vocabulary triples from a dataset of paper abstracts retrieved from the PubMed.

Although this open vocabulary approach can add complexity to the modeling of semantic relationships (and therefore to the learning problem), it also adds expressiveness to the resulting KBs. Such expressiveness can improve the contextual informativeness of semantic structures and thus allow the NLM to discriminate useful patterns from those that are not for the target reasoning task \cite{dhingra2020differentiable}. To show the effectiveness of our approach in the target reasoning task from multiple points of view, we evaluated and analyzed the results in terms of:

\begin{enumerate}[i.]
    \item \textbf{Performance metrics} in source tasks. Accuracy considers the reasoning quality as an exact prediction, with respect to the predicted and ground truth tokens. Cross Entropy quantify the degree of dissimilarity between the probability distribution predicted by the model and the ground truth one. These are effective at learning time because the inner representations of the NLM acquire knowledge on what are the specific words (the object phrase) that probably should be next to the input ones (the subject and predicate).
    \item \textbf{STS-based hipothesis testing.} When it comes to reasoning, specificity of performance metrics may not be characteristic of natural language, especially from the point of view of the open vocabulary inherent in natural language and semantic change (probably due to logical inference and/or synonymy). To account for these points of view,  we performed hypothesis testing based on Semantic Textual Similarity (STS) in source and target tasks, which measures reasoning quality in the sense of semantic relatedness \cite{arroyo2017lipn,arroyo2019unsupervised}. We compared to the distribution of STS measurements between predicted and shuffled ground truth object phrases (a random baseline that simulates perturbation of the actual correspondence between subject-predicate and object).
    \item \textbf{Inspection of the inferences} in the source and target tasks, which has the aim of showing the semantic regularities learned by the NLMs and how it holds from source to target tasks.
\end{enumerate}


Note that from the point of view of the ConceptNet KB as a source task, this has been mostly compiled by hand, and therefore, the task represents a distantly supervised learning problem. The target task represented by the OpenNCDKB was built unsupervised, so this task represents the target domain of a transfer learning problem.

\section{Theoretical Background}\label{sec:03_theoretical_background}

\subsection{Open Information Extraction for openKBs}
\label{sec:oie}
Open Information Extraction (OpenIE) is a set of IE methods that generalize semantic relations to open vocabulary semantic structures. The latter allow for a much broader coverage of the ideas expressed in natural language, even when the classic methods are rule-based \cite{etzioni2008open,fader2011identifying}.

OpenIE works in such a way that given the example sentence \textit{``habitat loss is recognized as the driving force in biodiversity loss''}, it generates semantic relations in the form of SPO triples, e.g. \{\texttt{
``habitat loss''}, \texttt{``is recognized as driving''}, \texttt{``biodiversity loss''}\}. In this triple, in the sense of Dependency Grammars \cite{10.2307/411934},\footnote{Unlike to Dependency Grammars, Phrase Structure Grammars (the classic approach), in that the predicate includes the verb and the object phrase. Instead, Dependency Grammars assumes the verbal form as the center (or highest hierarchy) of the sentence, linking the subject and the object. This latter is a convenient approach in building KBs, and therefore for OpenIE.} \texttt{
``habitat loss''} is the subject phrase, \texttt{``is recognized as driving''} is the predicate phrase, and \texttt{``biodiversity loss''} is the object phrase. In semantics, the subject is the thing that performs actions on the object, which is another thing affected by the action expressed in the predicate. Predicates, therefore, relate things (the subject and the object) in a directed way from the former to the latter.

OpenIE takes a sentence as input and outputs different versions of its SPO structure. Normally these versions are ``sub-SPO'' structures contained in the same sentence, e.g. \{\texttt{
``habitat loss''}, \texttt{``is recognized as''}, \texttt{``driving biodiversity''}\}, and \{\texttt{
``habitat loss''}, \texttt{``recognized as''}, \texttt{``driving''}\}. Notice that the last extraction (triplet) may not be factual at all and, depending on the downstream task this kind of output is used for, it may be considered purposeless.  

We use the obtained OpenIE triples to build an openKB that organizes knowledge from NCD-related paper abstracts (Section \ref{sec:open_ncd_kb_data}). In addition we used already existent extractions to build a general purpose openKB (with no specific topics, see Section \ref{sec:oie_gp_data}). A KB is a special case of a database that uses a structured schema to store both structured and unstructured data. In our case, the structured data constitute the identified elements of semantic triples, i.e. \{\texttt{subject, predicate, object}\}, while the unstructured part is the open vocabulary text (natural language phrases) of each of these elements. 

\subsection{Neural Semantic Reasoning Modeling}\label{sec:04_neural_reasoning_modeling}
Let $(s, v, o)\in\mathcal{K}$ be a semantic triple, where $s, v, o$ are subject, predicate and object phrases, respectively, and $\mathcal{K}$ is the training openKB. The KBC task here is to predict $o$ given $s,v$, which gives place to the conditional probability distribution $P(O|U_{s,v})$ implemented using a Neural Network model:
$$p(o|u)=f_{nn}(h_o,h_u),$$
\noindent where $O$ is the Random Variable (RV) that takes values on the set of object phrases $\mathcal{O}\ni o$, and $U_{s,v}$ is the RV that takes values on the set of concatenated subject-predicate phrases $\mathcal{U}\ni u=s\oplus v$. The Neural Network $f_{nn}(\cdot,\cdot)$ has learnable parameters $h_o,h_u\in\mathbb{R}^d$, which can be interpreted as phrase embeddings of $o$ and $u$, respectively. From the point of view of NMT, the probability mass function $p(o|u)$ can be used as a sequence prediction model. In this setting, each word $o_i$ of the target sequence (the object phrase) has a temporal dependency on prior words of the same phrase $o_{i'<i}$, and on the source sequence embedding $h_u$ (the concatenated subject-predicate phrases):
\begin{equation}
    \label{eq:seq_conditional}
    p(o_i|o_{i'<i}) = \sigma_d(o_{i-1},h_{i},h_{u}),
\end{equation}
\noindent where $\sigma_d(\cdot)$ is the decoder activation (a softmax function) that computes the probability of decoding the $i-$th (current) word of the object phrase from both, the current hidden state embedding $h_i$ and the prior state embedding $h_{i-1}$, as well as from the source embedding. Notice that in the case of modeling this sequence prediction problem using Recurrent Neural Networks (RNNs), the $i$ index represents time. Otherwise, it is simply the position of a word within the phrase.

\subsection{Transformers}


As first introduced for attention RNNs \cite{luong-etal-2015-effective,bahdanau2015neural}, we use the encoder-decoder Transformer as an NLM intended to generate object phrases (output) $o=o_1,\dots,o_n$ given the concatenated subject-predicate phrases (input) $u=s\oplus v=u_1,\dots,u_n$.\footnote{Notice that in this case we have sequences of the same length, $n$, in the input and in the output.} The input Transformer encoder block takes as input the $n$ $d-$dimensional (learnable) word embeddings of each item $u_i$ of the input sequence $u$ in parallel. Therefore, such input to the encoder is a matrix $X\in\mathbb{R}^{n\times d}$ (whose rows are word embeddings $x_i$) accepted by the $\ell-$th attention head of the $m-$headed multi-head self-attention layer \cite{dutta2021redesigning}:
\begin{equation}
    \label{eq:self_attention}
    H_{\ell}=\langle\sigma(\Lambda),W_v^\top X\rangle\rangle,
    \end{equation}
\noindent where $H_{\ell}\in\mathbb{R}^{n\times (d/m)}$ is the context matrix resulting from the $\ell-$th attention head, with $\ell=1,\dots,m$, and $\sigma(\cdot)$ is the element-wise softmax activation. The attention matrix $\Lambda\in\mathbb{R}^{n\times (d/m)}$ is given by
\begin{equation}
    \label{eq:self_attention_matrix}
    \Lambda=\frac{\langle W_q^\top X,W_k^\top X\rangle}{\sqrt{d/m}},
\end{equation}
\noindent where $W_q,W_k,W_v\in\mathbb{R}^{d\times (d/m)}$ are fully connected layers with linear activations (simple linear transformation layers), each entry $\alpha_{ij}\in\mathbb{R}$ of $\Lambda$ is the attention weight, from $u_i$ to each other $u_j$.

The multi-head self-attention layer builds its output by concatenating the $m$ context matrices:
$$H_{\oplus \ell}=\bigoplus_{\ell = 1}^m H_{\ell},$$
\noindent where then $H_{\oplus \ell}\in\mathbb{R}^{n\times m(d/m)}$ is fed to another linear output fully connected layer, i.e. $H'=W_{\oplus \ell}^\top H_{\oplus \ell}\in\mathbb{R}^{n\times d}$, whose output is in turn fed to the normalization layer given by:
$$f_N(z)=\gamma\frac{(z-\mu)}{\varsigma}+\beta,$$
\noindent where $f_N(\cdot)$ is applied both, to $X+H'$, therefore $H=f_N(X+H')$, and to the output of the block, i.e.  $H_b=f_N(H + W_b^\top H)\in\mathbb{R}^{h\times d}$, where $W_b\in\mathbb{R}^{h\times d}$ is an $h-$dimensional linear output fully connected layer ($h$ is the dimension of the latent space of the encoder-decoder model, i.e. the number of outputs of the encoder. In most cases $h=n$). The user defined parameters $\mu$ and $\varsigma$ of $f_N$ are the sample-wise mean and variance, i.e. over each input embedding of the layer.

To build the decoder, a second Transformer block is stacked to an input one, just after the first normalization layer of the latter (thus $W_b$ does not operate for the first block). This way, the output of the first decoder block is taken as the query of the second block, whose key and value are the output of the encoder. As in the case of any encoder-decoder configuration, the decoder takes the target sequence as input and output. The Transformer architecture allows the stacking of multiple blocks (layers), which also extends to the encoder and decoder. In this work we used $\{1,2\}-$block Transformer encoder-decoder NLMs.

\section{Data and Experimental Setup}\label{sec:05_data_and_experimental_setup}

\subsection{The OIE-GP Knowledge Base}\label{sec:oie_gp_data}
In this work, we created a dataset called OIE-GP by using manually annotated and artificially annotated OpenIE extractions. The main criterion for selecting the sources of the extractions was that they had some human annotation, either for the identification of the elements of the structure or for their factual validity. We considered factual validity as an important criterion because it is important to train the NLM to reason with factual validity.

\textbf{ClausIE} was used to generate a large amount of triples that were manually annotated according to their factual validity \cite{10.1145/2488388.2488420}.\footnote{\url{https://www.mpi-inf.mpg.de/departments/databases-and-information-systems/research/ambiverse-nlu/clausie}
} The resulting dataset provides annotations indicating if the triples are either too general or senseless (correct/incorrect), which recent datasets have adopted in some way. For example it is common to find negative samples (incorrect triples) like, e.g., \texttt{\{"he"; "states"; "such thing"\}}, \texttt{\{"he"; "states"; "he"\}}. From this dataset, we took the 3374 OpenIE triples annotated as correct (or positive) samples.

\textbf{MinIE-C}\footnote{MinIE-C is the less strict version of MinIE system, and we selected it as it is not restricted in the length of the slots of the triples: \url{https://github.com/uma-pi1/minie}} provided artificially annotated triples that are specially useful to our purposes because they are as natural as possible in the sense of open vocabulary \cite{gashteovski-etal-2017-minie}. These are generated by the ClausIE algorithm (as part of MinIE) and are annotated as positive/negative according to the same criterion used by ClausIE. From this dataset, we took the 33216 OpenIE triples annotated as positive samples. 

\textbf{CaRB} is a dataset of triples whose structure have been manually annotated (supervised) with n-ary relations \cite{bhardwaj-etal-2019-carb}. From this dataset, we took the 2235 triples annotated as positive samples. 

\textbf{WiRe57} contains supervised extractions along with anaphora resolution \cite{lechelle-etal-2019-wire57}. The 341 hand-made extractions of the dataset are 100\% useful as positive samples because they include anaphora resolution.

\subsection{The OpenNCDKB}\label{sec:open_ncd_kb_data}
A noncommunicable disease (NCD) is a medical condition or disease that is considered to be non-infectious. NCDs can refer to Chronic Diseases, which last long periods of time and progress slowly. We created a dataset of scientific paper abstracts related to nine different NCD domains: Breast Cancer, Lung Cancer, Prostate Cancer, Colorectal Cancer, Gastric Cancer, Cardiovascular Disease, Chronic Respiratory Diseases, Type 1 Diabetes Mellitus, and Type 2 Diabetes Mellitus. These are the most prevalent world-wide NCDs, according to the World Health Organization \cite{world2018noncommunicable}.



We used the names of the diseases as search terms to retrieve the $k\sim 150$ most relevant abstracts from the PubMed.\footnote{National Library of Medicine, National Center for Biotechnology Information (NCBI):  \url{https://pubmed.ncbi.nlm.nih.gov}} 
The resulting set of abstracts constituted our NCD dataset. To generate our Open vocabulary Chronic Disease Knowledge Base (OpenNCDKB), we retrieved a total of 1,200 article abstracts that correspond to the NCD-related domains mentioned above.

To obtain OpenIE triples (and therefore an open vocabulary KB), we used the CoreNLP and OpenIE-5 libraries \cite{manning2014stanford,saha-etal-2017-bootstrapping}. First, we took each abstract from the NCD dataset and split it into sentences using the coreNLP library. Afterwards, we took each sentence and extracted the corresponding OpenIE triples using the OpenIE-5 library. By doing so, we obtained a total of 22,776 triples.
These triples were filtered to remove the ones that contain only stop words in the subject or object phrases. After this preprocessing, we were left with 18,616 triples that were considered valid for our purposes. 



In addition to the valid triples, we also generated semantically incorrect negative samples (triples). These were generated using the same methods for Artificial Semantic Perturbations and were preprocessed in the same way as the positive ones, giving 45,032 semantically invalid triples. These negative samples can be used for training factual validity detection models, which resemble the plausibility score included in the ConceptNet Common Sense Knowledge Base \cite{li2016commonsense}.

\subsection{The Source Reasoning Tasks}\label{sec:source_tasks}
To perform training, we first used a compact version of the ConceptNet KG converted into a KB (the ConceptNet Common Sense Knowledge Base \cite{li2016commonsense}) \footnote{\url{https://home.ttic.edu/~kgimpel/commonsense.html}} consisting of 600k triples. We also used our OIE-GP Knowledge Base (Section \ref{sec:oie_gp_data}) containing 39166 triples, i.e. only the (semantically) positive samples of the whole data described in Section \ref{sec:05_data_and_experimental_setup}. 

From these KBs, we constructed mixed KBs that include the source task vocabulary and also include, as much possible, missing vocabulary needed to validate the model on the target task, related to NCDs. In this way, we obtained our source KBs:

\begin{enumerate}
    \item \textbf{OpenNCDKB}.
We split the 18.6k triples of the OpenNCDKB into 70\% (13.03k) for training, and 30\% for testing (5.58k). We included the OpenNCDKB here because in the context of source and target task, we simply split the whole KB into train, test and validation data. Validation data was considered the target task in this case.

    \item \textbf{ConceptNet+NCD}.
By merging the triples collected from the ConceptNet Knowledge Graph and those from OpenNCDKB, we obtained our ConceptNet+NCD KB containing 429.12k training triples and 183.91k test triples.

    \item \textbf{OIE-GP+NCD}.
We obtained our OIE-GP+NCD KB by merging 39.17k OIE-GP and 13.03k NCD triples to obtain a total of 52.20k triples. These were split into 70\% (36.54k) for training and 30\% (15.66k) for testing.

    \item \textbf{ConceptNet+OIE-GP+NCD}. We obtained this large KB that included ConceptNet, general purpose OpenIE (the OIE-GP KB) and NCD OpenIE triples (the OpenNCDKB) by taking the union between ConceptNet+NCD and OIE-GP+NCD. This source training task constitutes 600k + 39.17k + 13.03k = 652.20 total triples. These were split into 70\% (456.54k) for training and 30\% (195.66k) for testing.
\end{enumerate}

All the mentioned quantities consider that we filtered out the triples whose subject or object phrases were only stopwords.

\section{Results and Discussion}\label{sec:06_results_discussion}
\subsection{Experimental Setup}
For our experiments we used a previously validated Transformer encoder-decoder model to evaluate its performance on our open vocabulary KBC task. Due to the fact that our KBs are smaller than the datasets used for NMT by the authors of \cite{vaswani2017attention}, we decided to use the smallest model architecture they reported. The Base Model defined by these authors used a sequence length of $\max|u|=\max|o|=30$ (the maximum subject-predicate concatenation and object lengths in word-based tokens), a model dimension of $d=512$ (the input positional word embeddings), an output fully connected layer dimension of $h=2048$ (denoted $d_{ff}$ in the original paper), a number of attention heads of $m=8$, attention key and value embedding dimensions of $d/m=64$, and a number of transformer blocks of $N=2$. In addition, we considered the possibility that such ``small'' model is still too big for our KBs (the largest one has $\sim 652$k triples), compared to the 4.5 million sentence pairs this model consumed in the original paper for NMT tasks. Therefore, we also included an alternative version of the Base Model using only one Transformer block ($N=1$) in both the encoder and decoder,\footnote{In the case of the decoder, single Transformer block refers to two self-attention layers ($N=1$), whereas $N=2$ refers to three of these layers (i.e. the decoder's number of blocks is $N+1$ w.r.t the encoder).} while keeping all other hyperparameters.

The training of the models was performed using different KBs constructed from different sources to select the task  and the model that best transfers to our target task, the OpenNCDKB (see Section \ref{sec:source_tasks}). Using the four source KBs and the OpenNCDKB we obtained the four mixed KBs used for training: ConceptNet+NCD (CN+NCD, for short), IOE-GP+NCD, Concepnet+IOE-GP+NCD (CN+OIE-GP+NCD, for short). Using these, we obtained a total of eight models, four with $N=1$ and four with $N=2$.

We presented a comparison of the source task performance metrics of the eight models (training and test). We also performed an STS-based hypothesis test on the source and target tasks. STS is based on the cosine similarity between phrase embeddings. During the source tasks, and for each of the eight trained models, we first computed the STSs between the predicted and ground-truth object phrases (actual STS measurements). Then, we computed the STSs between the predicted phrases and a shuffled version of the ground-truth object phrases to obtain a random baseline that simulated the perturbation of the actual correspondence between subject-predicate and object. Therefore, hypothesis testing was first performed for the actual STS measurements and then for the randomized baseline. Next, the same computations were performed on the target tasks. The overall outcome of this experiment was to demonstrate whether the null hypothesis, i.e., that the means of the actual STS measurements and the random baselines come from the same distribution, could be rejected with confidence, and whether this held from the source to target tasks.

The neural word embeddings we used for STS-based hypothesis testing were trained on the Wikipedia corpus in order to obtain a good coverage of the set difference between the vocabulary of the source and target tasks (paper abstracts contain simpler vocabulary than the paper itself, while  Wikipedia contains a relatively technical vocabulary). The embedding algorithm used was FastText as it has showed better performance in representing short texts \cite{bojanowski2017enriching}. The phrase embedding method used to represent object phrases was simple embedding summation, this was because at the phrase level, even functional words (e.g. prepositions, copulative and auxiliary verbs) can change the meaning of the represented linguistic sample \cite{arroyo2019unsupervised}.

Finally, we analyzed by inspection the natural language predictions of the best models with N=1 and N=2, in both source and target tasks. For this purpose, five input samples were randomly chosen from the source tasks, and the NLMs that showed the highest confidence during our STS hypothesis testing were fed with them. The obtained predictions were analyzed from the point of view of their meaning, and semantic regularities they showed. Next, in the target task (OpenNCDKB), we performed the same inspection on a random sample of five subject-predicate inputs not seen by the trained models. This was to verify whether the semantic regularities previously identified in the source task predictions were reproduced in the target task.

\subsection{Results}

\subsubsection{Performance Metrics}

\begin{figure}[htb]
	\centering
	\hspace*{\fill}
	\begin{subfigure}[b]{.48\linewidth}
		\centering
		\includegraphics[width=\linewidth]{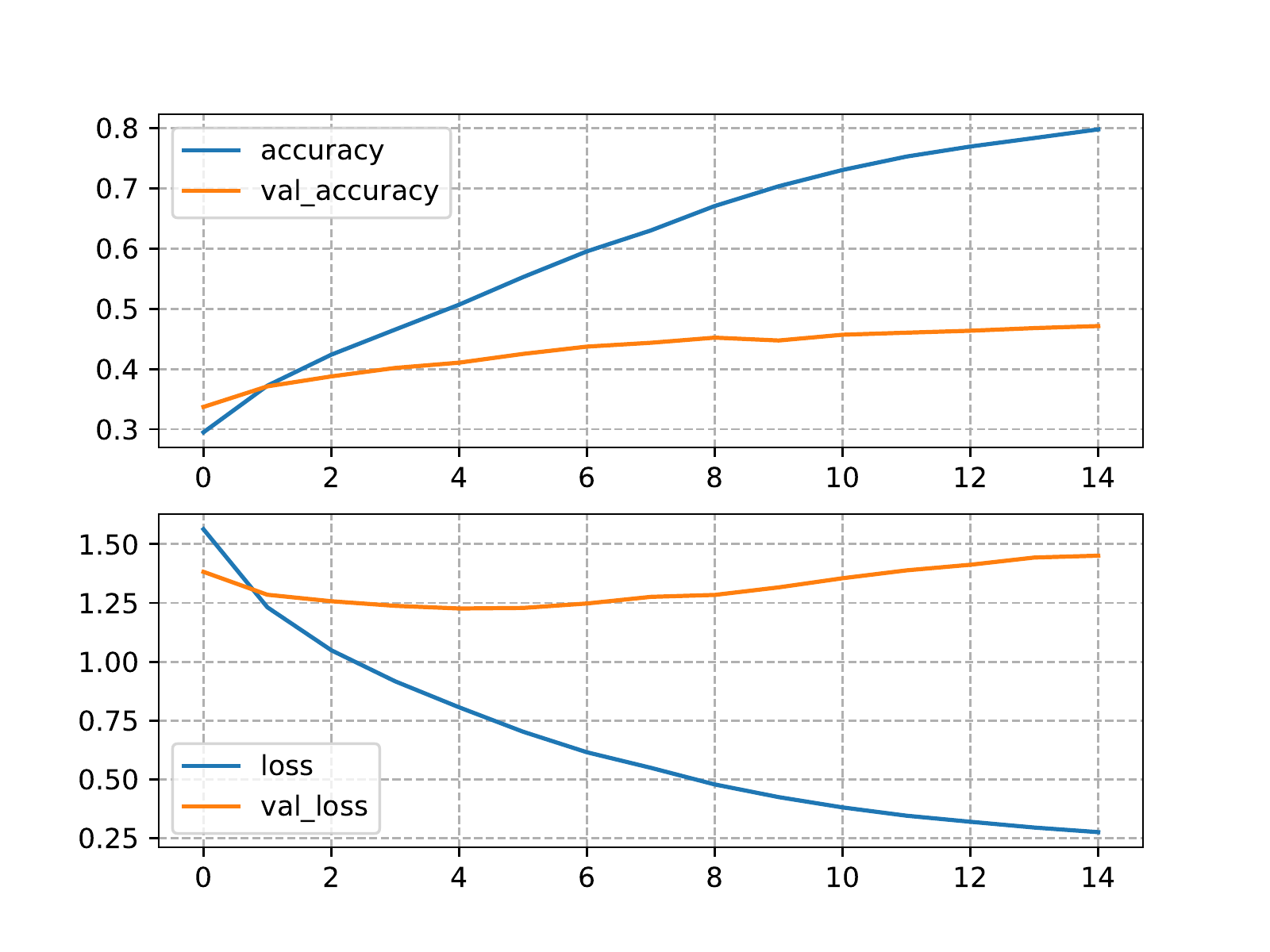}
		\captionsetup{justification=centering}
		\caption{}
         \label{fig:ncd_hist_plot_N1}
	\end{subfigure}
	\hspace*{\fill}
	\begin{subfigure}[b]{.48\linewidth}
		\centering
		\includegraphics[width=\linewidth]{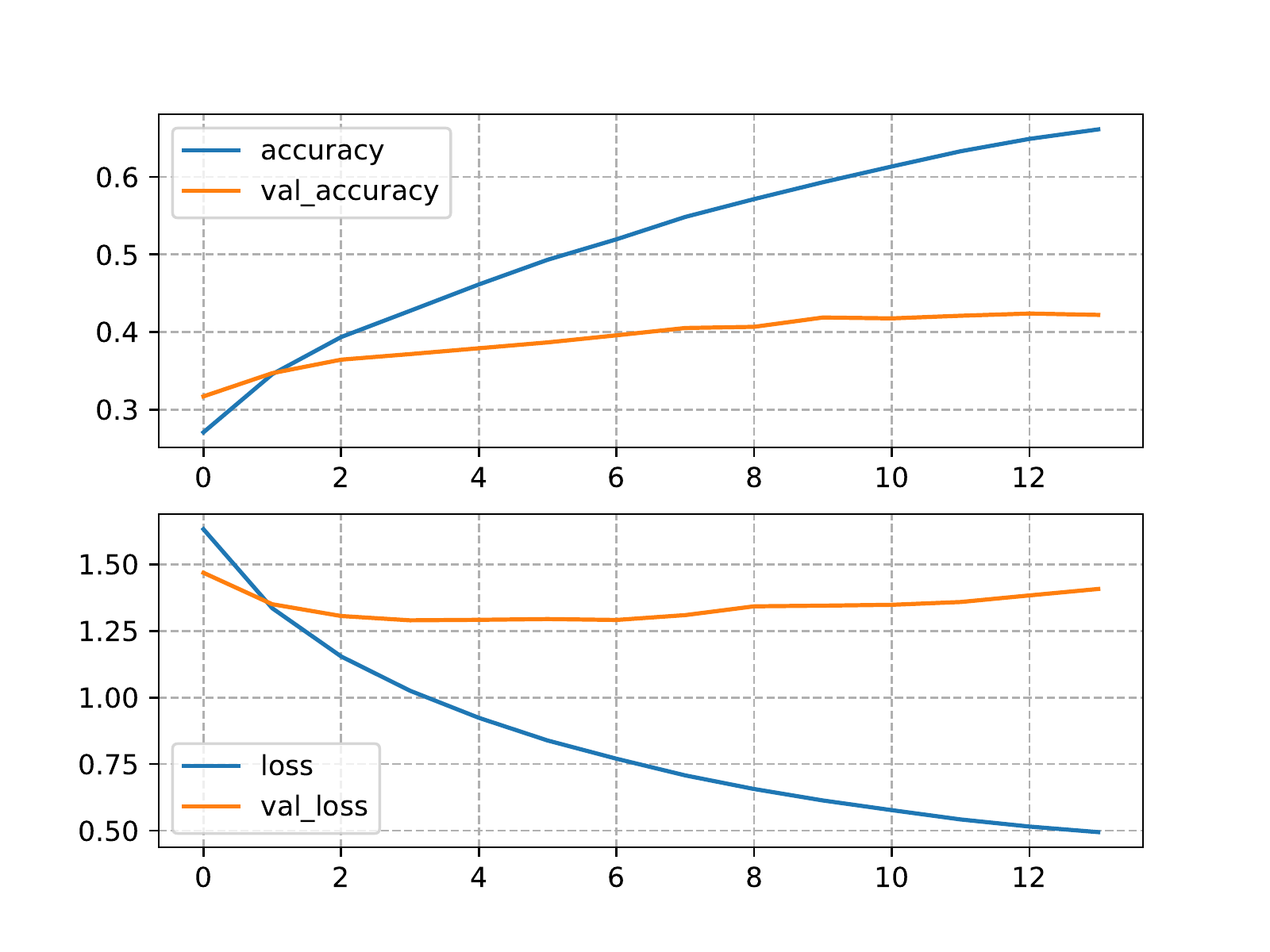}
		\captionsetup{justification=centering}
		\caption{}
         \label{fig:ncd_hist_plot_N2}
	\end{subfigure}
	\hspace*{\fill}
	\caption{Training and test (\texttt{val\_}) accuracies (upper plots), and training and test (\texttt{val\_}) losses (lower plots) for the Transformer trained with the OpenNCDKB: (\textbf{a}) N=1, (\textbf{b}) N=2.}
	\label{fig:ncd_hist_plot}
\end{figure}

In Figure \ref{fig:ncd_hist_plot}, we show the progress of the performance metrics of the Transformers trained and tested only on the OpenNCDKB. Figure \ref{fig:ncd_hist_plot_N1} shows the accuracy and the loss for the one-block Transformer NLM. Notice that the difference between train and test accuracy (\texttt{accuracy}=80\% and \texttt{val\_accuracy}=48\%, respectively) is relatively large. The same occurred for the loss function (\texttt{loss}=1.28bits and \texttt{val\_loss}=0.25bits). This can be the manifestation of overfitting, although the NLM showed to be relatively stable through the fourteen epochs allowed by the patience hyperparameter.

\begin{figure}[htb]
	\centering
	\hspace*{\fill}
	\begin{subfigure}[b]{.48\linewidth}
		\centering
		\includegraphics[width=\linewidth]{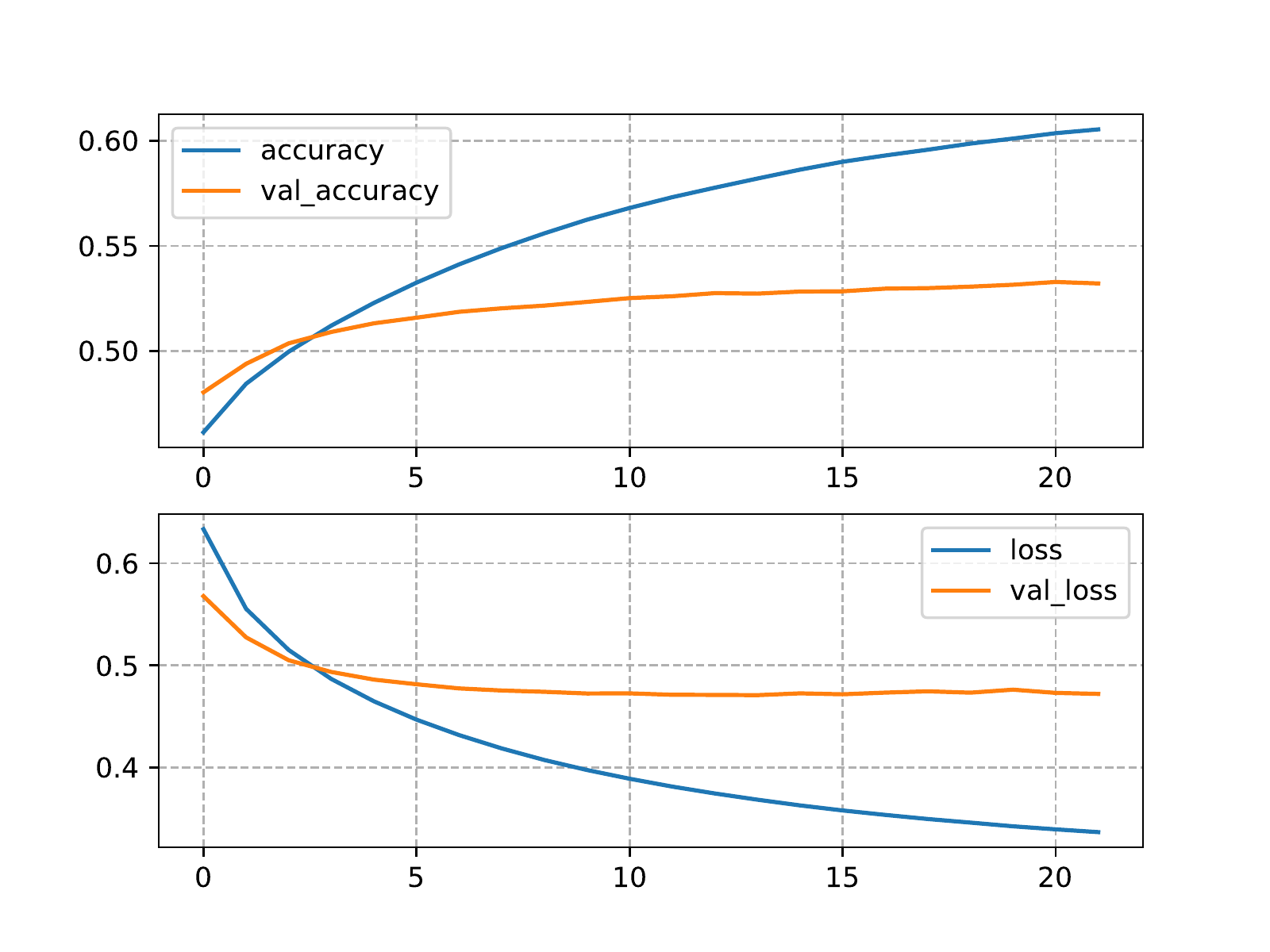}
		\captionsetup{justification=centering}
		\caption{}
        \label{fig:ncd_conceptnet_hist_plot_N1}
	\end{subfigure}
	\hspace*{\fill}
	\begin{subfigure}[b]{.48\linewidth}
		\centering
		\includegraphics[width=\linewidth]{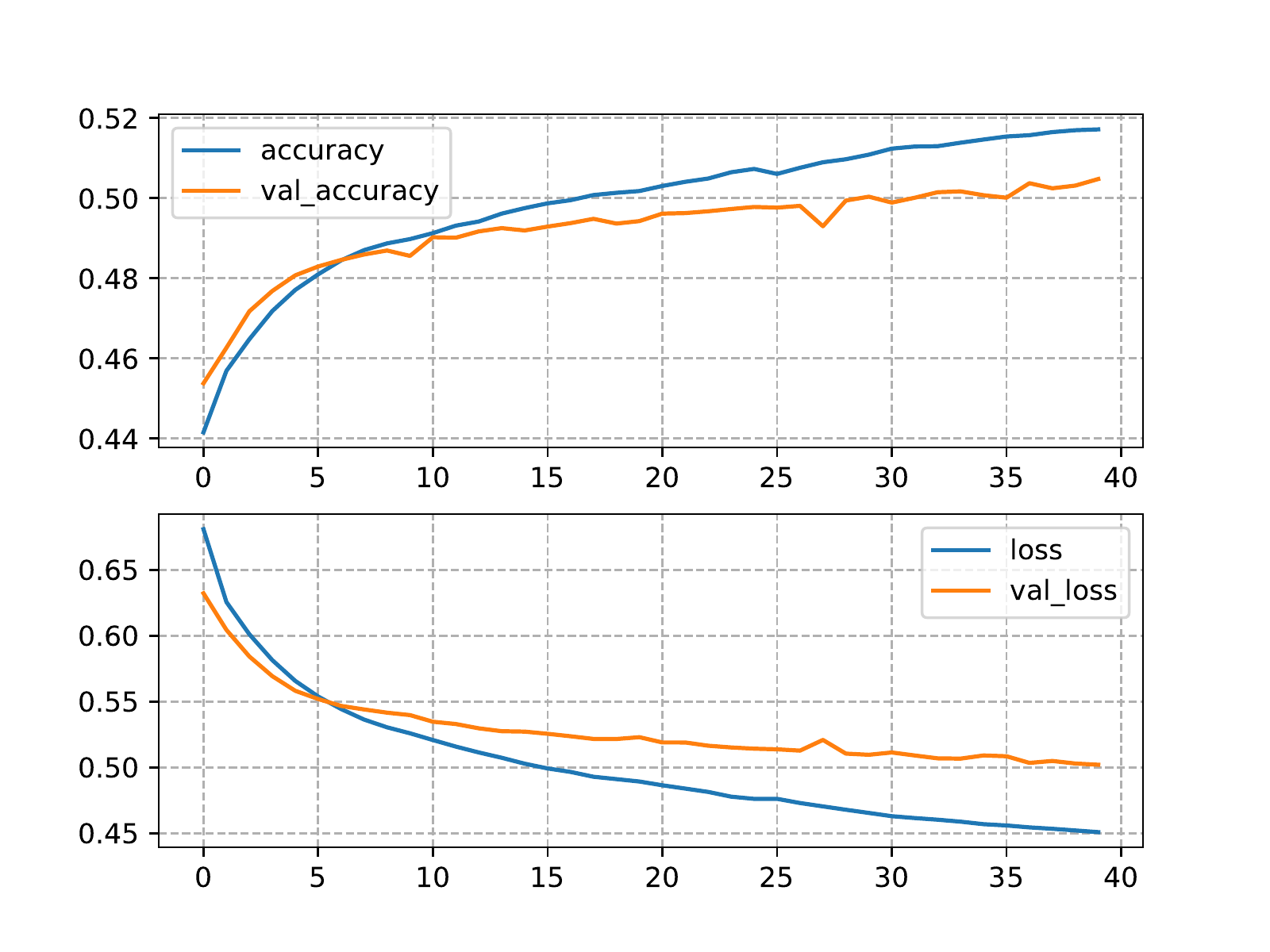}
		\captionsetup{justification=centering}
		\caption{}
        \label{fig:ncd_conceptnet_hist_plot_N2}
	\end{subfigure}
	\hspace*{\fill}
	\caption{Training and test (\texttt{val\_}) accuracies (upper plots), and training and test (\texttt{val\_}) losses (lower plots) for the Transformer trained with the CN+NCD KB: (\textbf{a}) N=1, (\textbf{b}) N=2.}
	\label{fig:ncd_conceptnet_hist_plot}
\end{figure}

Similarly to the one-block model, the accuracy and the loss (Figure \ref{fig:ncd_hist_plot_N2}) for the two-block Trasformer NLM shows a manifestation of overfitting. Also, the two-block NLM converged two epochs earlier than the one-block NLM.
The models trained with the CN+NCD KB were provided with much more data than the models only using the OpenNCDKB. The accuracy and loss for these models are shown in Figure \ref{fig:ncd_conceptnet_hist_plot}. Figure \ref{fig:ncd_conceptnet_hist_plot_N1} shows a clear improvement in the test loss, reaching a minimum near to 0.48bits. On the other hand, the training loss was about 0.13bits apart from the test loss, which is less than the same comparison made for the two-block NLM trained only with the OpenNCDKB. Regarding accuracy, the maximum on test data was about 53\%, showing similar train and test curves with respect to OpenNCDKB.

In the case of the two-block NLM trained with the CN+NCD KB (Figure \ref{fig:ncd_conceptnet_hist_plot_N2}), the training and test losses developed with lower divergence, the same can be said for the accuracies, possibly indicating stability and low variability of the model when it is exposed to unseen data (generalization). In addition, the test loss reached a minimum of 0.5bits, and did not diverge as much as in the one-block NLM. The maximum test accuracy surpassed 50\%, and the training accuracy behaved in a similar way. It is worth noting that the model improved throughout the 40 epochs we set as maximum, which suggests that the results could be improved by increasing the total number of epochs.

\begin{figure}[htb]
	\centering
	\hspace*{\fill}
	\begin{subfigure}[b]{.48\linewidth}
		\centering
		\includegraphics[width=\linewidth]{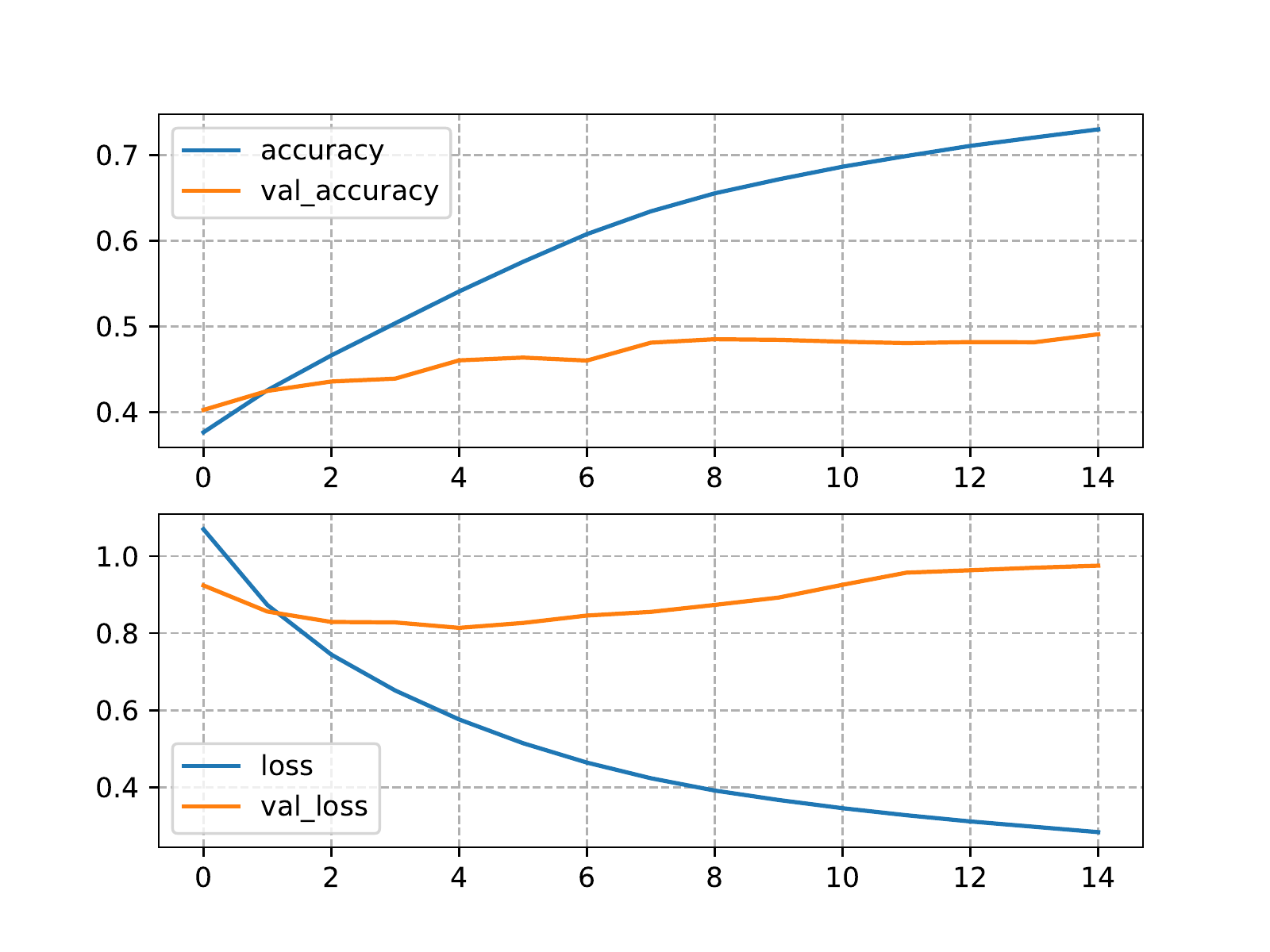}
		\captionsetup{justification=centering}
		\caption{}
        \label{fig:ncd_gp_hist_plot_N1}
	\end{subfigure}
	\hspace*{\fill}
	\begin{subfigure}[b]{.48\linewidth}
		\centering
		\includegraphics[width=\linewidth]{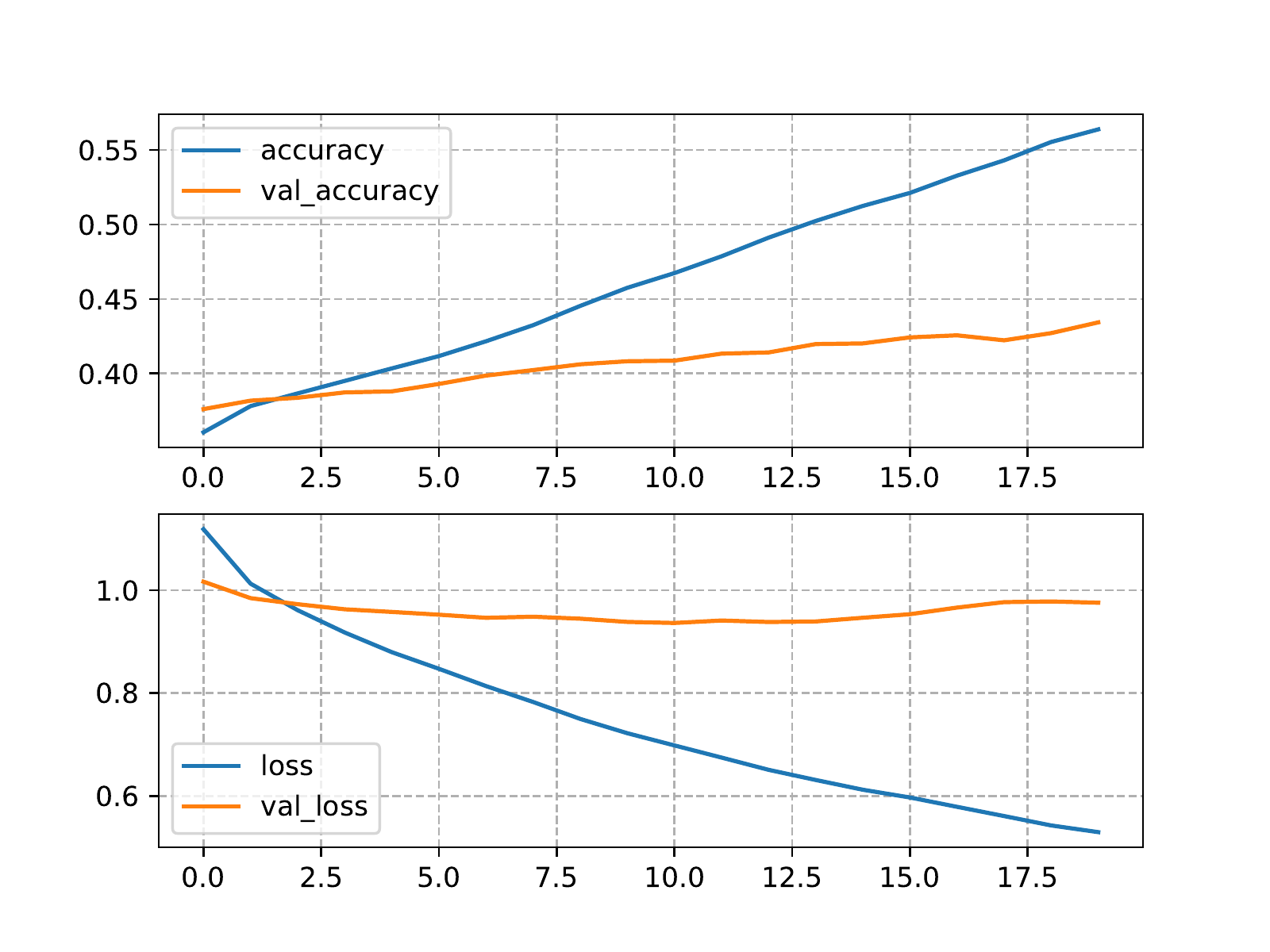}
		\captionsetup{justification=centering}
		\caption{}
        \label{fig:ncd_gp_hist_plot_N2}
	\end{subfigure}
	\hspace*{\fill}
	\caption{Training and test (\texttt{val\_}) accuracies (upper plots), and training and test (\texttt{val\_}) losses (lower plots) for the Transformer trained with the OIE-GP+NCD KB: (\textbf{a}) N=1, (\textbf{b}) N=2. }
	\label{fig:ncd_gp_hist_plot}
\end{figure}

The apparent lack of data of our smaller KBs was again highlighted by the results obtained from the NLM trained with the OIE-GP+NCD KB (see Figure \ref{fig:ncd_gp_hist_plot}). Although the OIE-GP KB is twice as large as OpenNCDKB, its barely one tenth the size of ConceptNet. The losses and accuracies shown in Figures \ref{fig:ncd_gp_hist_plot_N1} and \ref{fig:ncd_gp_hist_plot_N2} indicate that there are no clear improvements of using this KB over the OpenNCDKB.

\begin{figure}[htb]
	\centering
	\hspace*{\fill}
	\begin{subfigure}[b]{.48\linewidth}
		\centering
		\includegraphics[width=\linewidth]{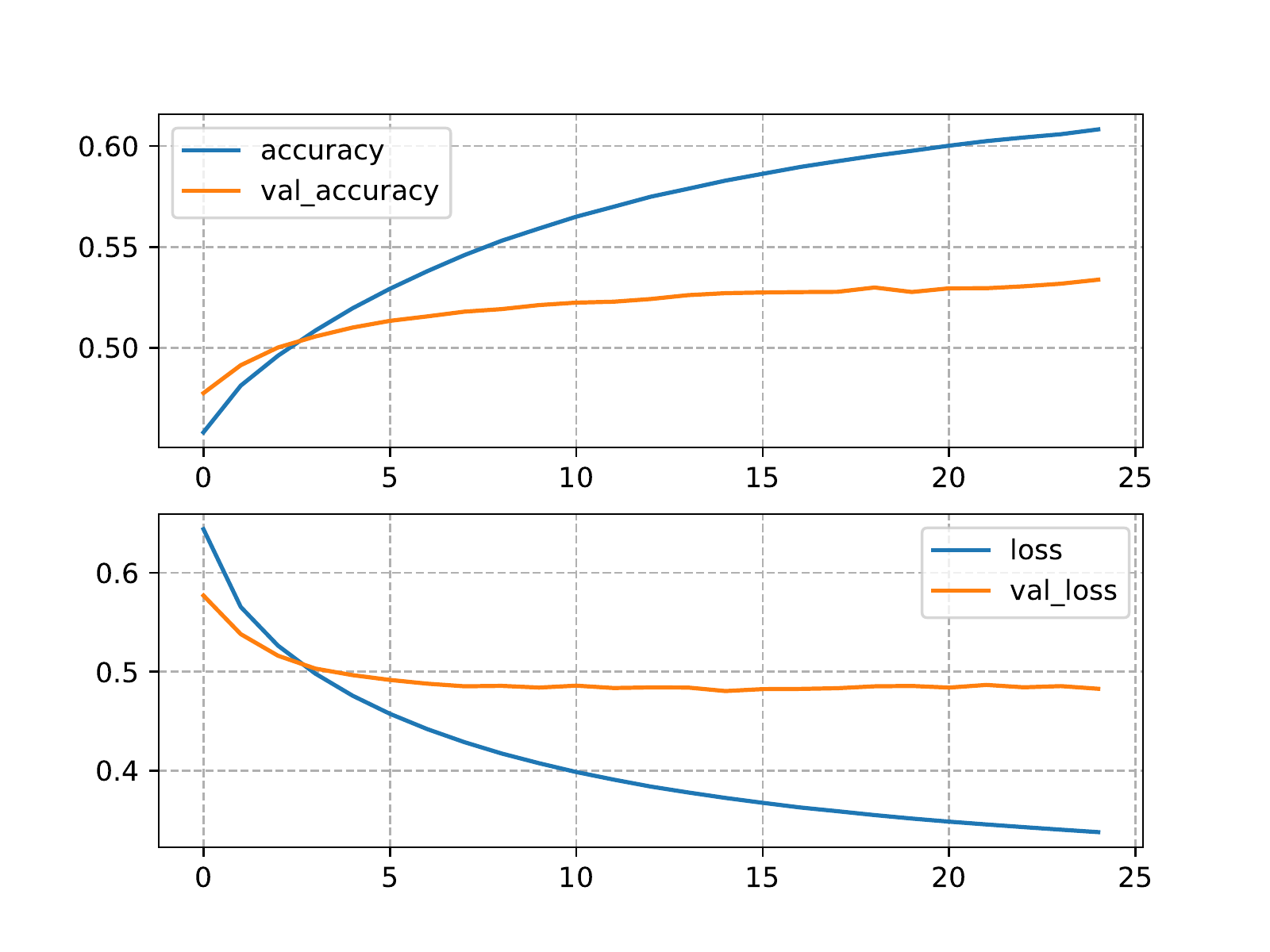}
		\captionsetup{justification=centering}
		\caption{}
        \label{fig:ncd_gp_conceptnet_hist_plot_N1}
	\end{subfigure}
	\hspace*{\fill}
	\begin{subfigure}[b]{.48\linewidth}
		\centering
		\includegraphics[width=\linewidth]{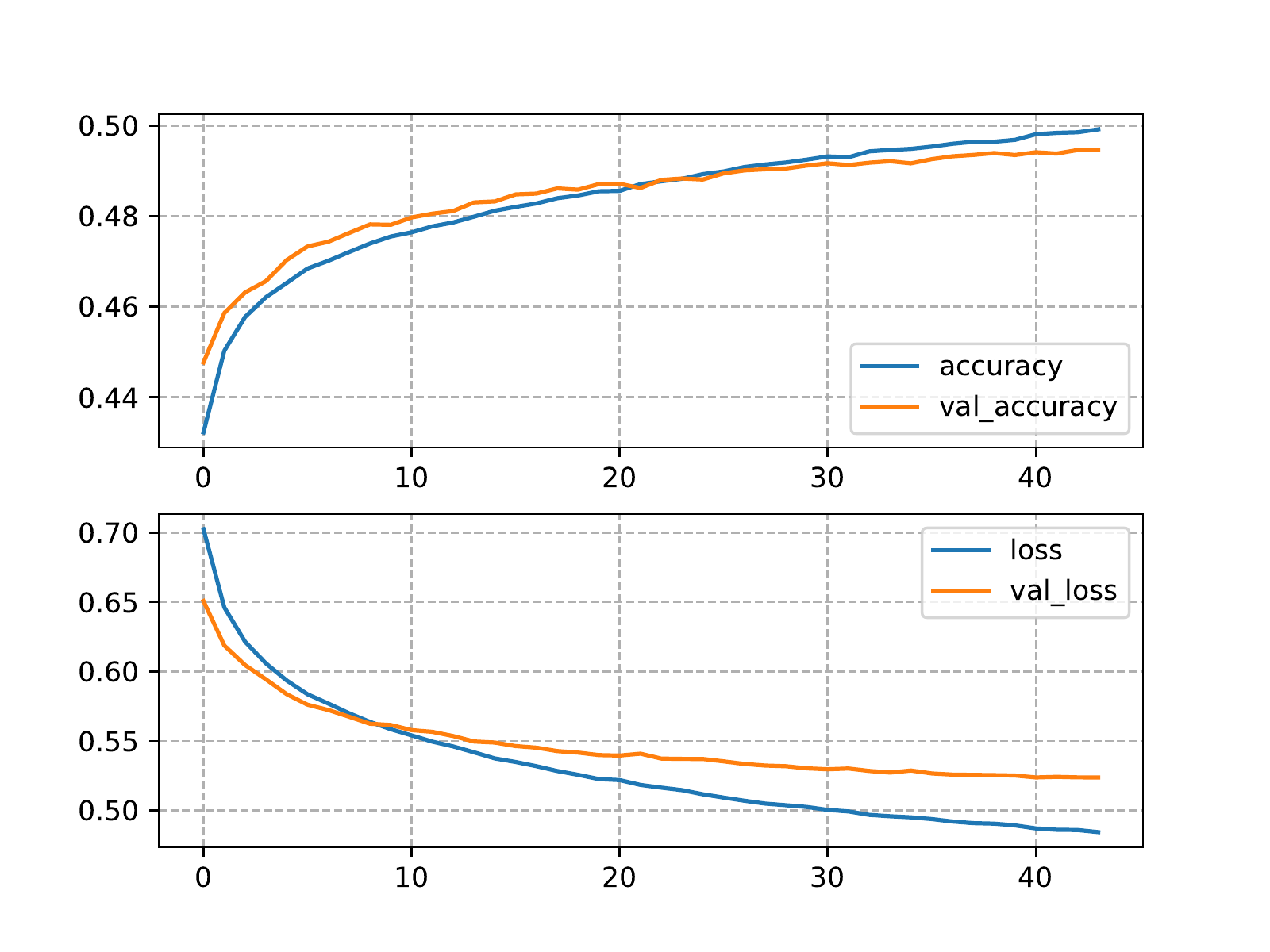}
		\captionsetup{justification=centering}
		\caption{}
        \label{fig:ncd_gp_conceptnet_hist_plot_N2}
	\end{subfigure}
	\hspace*{\fill}
	\caption{Training and test (\texttt{val\_}) accuracies (upper plots), and training and test (\texttt{val\_}) losses (lower plots) for the Transformer trained with the ConceptNet+OIE-GP+NCD KB: (\textbf{a}) N=1, (\textbf{b}) N=2.}
	\label{fig:ncd_gp_conceptnet_hist_plot}
\end{figure}

Finally, the NLMs trained using the ConceptNet+OIE-GP+NCD KB show similar results to the ones obtained with the ConceptNet+NCD KB (see Figure \ref{fig:ncd_gp_conceptnet_hist_plot}). The one-block NLM (Figure \ref{fig:ncd_gp_conceptnet_hist_plot_N1}) required two less epochs, compared to the same architecture trained on the ConceptNet+NCD KB. Training the two-block NLM (Figure \ref{fig:ncd_gp_conceptnet_hist_plot_N2}) with the ConceptNet+OIE-GP+NCD KB caused a decrease in performance, compared to training the model with the ConceptNet+NCD KB, i.e. 0.03bits in the case of the loss and 0.01\% in the case of accuracy. Despite this decrease, the model showed a greater stability, as the train and test curves were less divergent when the ConceptNet+OIE-GP+NCD KB was used.

\subsubsection{Semantic Textual Similarity}
We compared predictions with respect to random baselines using STSs between the predicted and ground truth object phrases (\texttt{pred}), and between the predicted and randomized ground truth object phrases (\texttt{rdn}). See Table \ref{tab:source_and_target_pvalues}. Our most significant model resulted when using the CN+NCD KB for training (the null hypothesis can be safely rejected). Its mean similarity with respect to the ground truth in the source task ($\mu$-Src \texttt{pred}) was 0.551 ($p\sim 0.0$), while in the target task ($\mu$-Trg \texttt{pred}) it was 0.414 ($p\sim 1.35\times 10^{-25}$). Such model required one Transformer block (encoder/decoder) and 20 training epochs. The performance of this model was followed by the NLM constituted by one-block Transformer trained with the CN+OIE-GP+NCD KB ($p\sim 1.32\times 10^{-22}$ in the target task).
\begin{table}
\begin{tabular}{l|l|l|l|l|l}
\hline
N & KB & $\mu$-Src (pred/rdn) & Source $p$ & $\mu$-Trg (pred/rdn) & Target $p$ \\
\hline
1 & CN+OIE-GP+NCD & 0.506 / 0.468 & 0.0 & 0.414 / 0.309 & $1.32\times 10^{-22}$ \\
\hline
1 & CN+NCD & 0.551 / 0.511 & 0.0 & 0.414 / 0.315 & $\mathbf{1.35\times 10^{-25}}$ \\
\hline
1 & OIE-GP+NCD & 0.584 / 0.569 & $5.16\times 10^{-179}$ & 0.459 / 0.404 & $1.95\times 10^{-8}$ \\
\hline
1 & OpenNCDKB & 0.613 / 0.596 & $3.31\times 10^{-7}$ & 0.610 / 0.595 & $2.83\times 10^{-11}$ \\
\hline
2 & CN+OIE-GP+NCD & 0.424 / 0.418 & 0.0 & 0.342 / 0.298 & 0.0708 \\
\hline
2 & CN+NCD & 0.500 / 0.492 & 0.0 & 0.367 / 0.310 & \textbf{0.0468} \\
\hline
2 & OIE-GP+NCD & 0.308 / 0.308 & 1.0 & 0.284 / 0.284 & 1.0 \\
\hline
2 & OpenNCDKB & 0.607 / 0.607 & 1.0 & 0.604 / 0.604 & 1.0 \\ 
\end{tabular}
\caption{P-values obtained using the source and target test KBs (bold p-values indicate the most confident models with N=1 and N=2, respectively).}
    \label{tab:source_and_target_pvalues}
\end{table}

In the case of the two-block NLMs, most of their predictions on the NCD-related target task can be confidently regarded as random (the null hypothesis holds), even those of the model trained with our largest KB (CN+OIE-GP+NCD). Only the two-block model trained with CN+NCD proved to be significantly reliable, but with relatively low mean STS on the target task ($\mu$-Trg \texttt{pred} of 0.367, $p=0.0468<0.05$). Based on our performance analysis, in which we had observed that the two-block models generalized well, but underperformed on the source task ($p\sim 0.0$), we believe that this low confidence of the STS measurements in the target predictions may be due to an overfitting of the models to the task level on the source task, which in turn can be attributed to model size. Despite this, the predictions of the one-block models trained with CN+OIE-GP+NCD and CN+NCD were the most confident in distinguishing from the random baseline in the source task. The only two-block model that resulted significant barely reached $p<0.05$, which was trained with the CN+NCD KB. These models trained with the smaller KBs (OIE-GP+NCD and OpenNCDKB) evidenced their poor performance (which initially appeared to be generalization), as their predictions can be reliably discarded as random in both the source and target task ($p\sim 1.0$).

Overall, according to our STS-based hypothesis tests, all source tasks provided the needed knowledge to the one-block NLMs to consistently and confidently be distinguished from the random baseline in both source and target tasks. This means that the models did not generate random meanings for the object phrases and that the generated meanings are in fact correlated with those of the ground truth. 

Notice also that most differences between the actual STSs and the random baselines (\texttt{pred-rdn}) were relatively small and the maximum resulted from the second most significant model (the one-block NLM trained with CN+OIE-GP+NCD), i.e. $0.414 - 0.309 \Rightarrow 10.5\%$. However, very small p-values also indicated low variability and high isolation between randomness and predictions, and thus high stability in the decisions the model made to generate meanings.

\subsubsection{Inspection of Predictions}

In Table \ref{tab:CN_source_tests_inferences} we show the predictions for five subject-predicate concatenations randomly sampled from the ConceptNet KB. In the case of \textit{``music is a''} (1), the NLMs predict, on the one hand, the exact object phrase \textit{``a form of communication''} (CN+OIE-GP+NCD), and on the other hand it predicted a synonymous phrase, \textit{``a form of expression''} (CN+NCD, N=1 and N=2), which sounds even more appropriate considering the context (arts), i.e. replacing \textit{``communication''} by \textit{``expression''}. The predictions of the one-block NLMs for the second input (2) can be seen as an ``objectification'' of the ground truth. The models replace the subjective idea of \textit{``enjoying''} an activity (riding a bicycle) with the objective, and therefore simpler, idea of the purpose for which the activity is conceived (its functional definition). In this case, there is no logical implication from the ground truth, but from the subject-predicate to the objective benefits that someone can obtain by doing the activity with respect to not doing it. The two-block NLM, however, tried to learn subjectivity but with an imprecise semantic change (\textit{``you like to play''}). A similar pattern could be observed again in the following sample (3), \textit{``car receives action''}, which focused on the action that cars are most likely to receive: being driven, rather than being propelled. This is called predicate-subject semantic agreement. We believe that these two interesting semantic regularities that the NLM learned can most likely be thought of as common sense (what the source task taught the model), rather than what the ground truth indicated. The two-block model attempted to attribute an abstract application to cars, rather than paying attention to the semantic and grammatical agreements between predicate and subject (\textit{``receives action''}). 

The case of the next sample (4), \textit{``an Indian restaurant used for''}, the NLMs to exhibit the above regularities and also show an additional manifestation of performance generalization in language generation. The input specifies the type of restaurant, and implicitly the country of origin of the food it sells, but the NLM probably could not learn to pay attention to such specific information. At the level of semantics, these manifestations of generalization of the model can also explain its preference for the more probable arguments of the predicate over the more specific ones in (3).
\begin{table}
\resizebox{14cm}{!}{%
\begin{tabular}{l|p{6cm}|p{10cm}}
\hline
\multirow{3}{*}{1} &\multicolumn{2}{l}{ music is a [start] form of communication [end] }\\\cline{2-2}
& CN+OIE-GP+NCD &  [start] a form of communication [end] \\
& CN+NCD &  [start] a form of expression [end] \\
& CN+NCD (N=2) &  [start] a form of expression [end] \\
\hline
\multirow{3}{*}{2} &\multicolumn{2}{l}{ ride a bicycle motivated by goal [start] enjoy riding a bicycle [end] }\\\cline{2-2}
& CN+OIE-GP+NCD &  [start] you are tired [end] \\
& CN+NCD &  [start] you need to get somewhere [end] \\
& CN+NCD (N=2) &  [start] you like to play [end] \\
\hline
\multirow{3}{*}{3} &\multicolumn{2}{l}{ car receives action [start] propel by gasoline or electricity [end] }\\\cline{2-2}
& CN+OIE-GP+NCD &  [start] drive by car [end] \\
& CN+NCD &  [start] drive by car [end] \\
& CN+NCD (N=2) &  [start] find in car [end] \\
\hline
\multirow{3}{*}{4} &\multicolumn{2}{l}{ an indian restaurant used for [start] selling indian meals [end] }\\\cline{2-2}
& CN+OIE-GP+NCD &  [start] eat a meal [end] \\
& CN+NCD &  [start] eating [end] \\
& CN+NCD (N=2) &  [start] eating [end] \\
\hline
\multirow{3}{*}{5} &\multicolumn{2}{l}{ mediators capable of [start] settle a disagreement [end] }\\\cline{2-2}
& CN+OIE-GP+NCD &  [start] eat cat food [end] \\
& CN+NCD &  [start] be [UNK] [end] \\
& CN+NCD (N=2) &  [start] be [UNK] [end] \\
\end{tabular}}
\caption{Five samples taken from the ConceptNet test dataset and used for inference using the specified models.}
    \label{tab:CN_source_tests_inferences}
\end{table}
 
The case of entry number five was atypical with respect to the objectivity that the NLMs try to induce in their predictions so far. From a general point of view of the meaning of the word ``mediator'', there is no relation to cats (see the prediction made by the NLM trained on the CN+OIE-GP+NCD KB). However, there is a sense of that word related to animal welfare organizations, for which ``mediators'' are representatives who are concerned with the prevention of animal abuse. So, was it just the addition of some general purpose knowledge (OIE-GP) to CN+NCD that allowed the NLM to infer a (rare) capability for ``mediators''? And, therefore allowing the (rare) prediction of \textit{``mediators eat cat food''}. The model trained with the KB CN+NCD (N=1 and N=2) did not find things that mediators can do (so the \texttt{[UNK]} output token). Note that, in general, the random samples analyzed so far showed that the two-block models did not learn semantic agreement between the elements of the semantic structures, even when they managed to distinguish themselves from the random baseline in the source task. This suggests that to be more confident about the knowledge acquired by the NLM, it must pass the STS-based hypothesis test with high significance in both the source and target task (from CSR to an arbitrary but specialized domain KBC, in our case).
\begin{table}
\resizebox{14cm}{!}{%
\begin{tabular}{l|p{5cm}|p{10cm}}
\hline
\multirow{3}{*}{1} &\multicolumn{2}{l}{ these subtypes are closely correlated [start] to clinical variables [end] }\\\cline{2-2}
& CN+OIE-GP+NCD &  [start] with a [UNK] [end] \\
& CN+NCD &  [start] with cardiovascular disease [end] \\
& CN+NCD (N=2) &  [start]with the development of the management of the disease[end] \\
\hline
\multirow{3}{*}{2} &\multicolumn{2}{l}{ Immune - based interventions are opening up [start] entirely novel therapeutic avenues [end] }\\\cline{2-2}
& CN+OIE-GP+NCD &  [start] to treat the disease [end] \\
& CN+NCD &  [start] [UNK] to cure individuals [end] \\
& CN+NCD (N=2) &  [start] to the development of the disease [end] \\
\hline
\multirow{3}{*}{3} &\multicolumn{2}{l}{ glycemic targets should reflect [start] presence of comorbid conditions [end] }\\\cline{2-2}
& CN+OIE-GP+NCD &  [start] the risk of gastric cancer [end] \\
& CN+NCD &  [start] the risk of microvascular complications such as retinopathy in patients with type 2 diabetes mellitus [end] \\
& CN+NCD (N=2) &  [start] the [UNK] of the [UNK] of the [UNK] [end] \\
\hline
\multirow{3}{*}{4} &\multicolumn{2}{l}{ these factors may provide [start] an opportunity to reduce deaths [end] }\\\cline{2-2}
& CN+OIE-GP+NCD &  [start] the risk of lung cancer [end] \\
& CN+NCD &  [start] more effective in treatment options in lung cancer [end] \\
& CN+NCD (N=2) &  [start] the [UNK] of the [UNK] [end] \\
\hline
\multirow{3}{*}{5} &\multicolumn{2}{l}{ Ethnic minorities to have [start] worse control of their diabetes [end] }\\\cline{2-2}
& CN+OIE-GP+NCD &  [start] the risk of lung cancer [end] \\
& CN+NCD &  [start] more likely to be high among patients [end] \\
& CN+NCD (N=2) &  [start] the treatment of gastric cancer [end] \\
\end{tabular}}
\caption{Five samples randomly chosen from the inferences made over the OpenNCDKB test set using the most significant models.}
    \label{tab:ncd_valid_inferences}
\end{table}

The best models obtained from source tasks were used to obtain predictions on validation samples obtained from the OpenNCDKB as a target task. To compare these models using these samples, we performed predictions on 5 randomly selected triples and listed them in Table \ref{tab:ncd_valid_inferences}.
The case of the triples randomly selected from the OpenNCDKB showed some very general inputs. This is the case of the inputs (1) and (4). For example, in the case of the model trained with CN+NCD (N=1), it gives what may be the most likely, grammatically well-formed prediction: ``cardivascular disease'' given ``subtypes closely correlated'', which is not factual at all as there is no a concrete subject in the input. The two-block model generated much more free constructions in this case.

The second input, \textit{``Immune-based interventions are opening up''} (2), causes the models to reason with very similar but much simpler meanings (\textit{``to treat the disease''}, and \textit{``to cure individuals''}) than the ground truth (\textit{``entirely novel therapeutic avenues''}). This was a reasoning behavior consistently observed during the inspection of the source task.
In the case of the input \textit{``glycemic targets should reflect''} (3) provides a concrete subject (\textit{``glycemic targets''}) which is related to \textit{``comorbid conditions''} in the ground truth sentence. Here the model (trained with the CN+OIE-GP+NCD KB) proposed that \textit{``glycemic targets''} also should be reflected in \textit{``the risk of gastric cancer''}, which can be verified as a valid fact in \cite{augustin2004glycemic}. The prediction made by the model trained with the CN+NCD KB could also be confirmed in \cite{lachin2008effect}. Last, the fifth input \textit{``Ethnic minorities to have''} (5), is interesting because the models seemed to learn more abstract meaning than ground truth. While the ground truth focuses on a specific disease (\textit{``Ethnic minorities to have worse control of their diabetes''}), the models reason that this is not only true for diabetes, but also for lung cancer, and that in general any sick person is more likely to be a minority.

Since the OpenNCDKB was made from abstracts, we had the opportunity to see some sentences that are not so specialized at all. This allowed us to confirm the replication of the behavior shown by NLMs from source to the target NCD-related task. This statement may be more difficult to make for the case of more specific knowledge. For example, it might require specialized boimedical knowledge to know whether \textit{the risk of gastric cancer} and \textit{the risk microvascular complications} are more likely and concrete conditions that can be affected or controlled by \textit{glycemic targets} than any \textit{comorbid condition}. 




\section{Related Work}\label{sec:07_related_work}
Regarding systems with IE, and semantic reasoning capabilities, we found common traits shared with the seminal work in Chronic Disease KGs \cite{shi2017semantic}. The authors proposed a data model to organize and integrate the knowledge extracted from text into graphs (ontologies, in fact), and a set of rules to perform reasoning via first-order predicate logic over a predefined dictionary of entities and relations \cite{bizon2019robokop}. More recently, \cite{li2020real} proposed association rule learning for relation extraction for KG construction, and neural network-based graph embedding for entity clustering from EMRs. In \cite{8758152}, the authors constructed a KG of gene-disease interactions from the literature on co-morbid diseases. They predicted new interactions using embeddings obtained from a tensor decomposition method. The authors of \cite{du2020knowledge} proposed a KG of drug combinations used to treat cancer, which was built from OpenIE triples filtered using different thesaurus \cite{RINDFLESCH2003462,Wei2013PubTatorAW}. The drug combinations were inferred directly from the co-occurrence of different individual drugs with fixed predicate and disease. The authors created their resource from the conclusions of clinical trial reports and clinical practice guidelines related with antineoplastic agents.

An EMR-based KG was used as part of a feature selection method for a support vector machine to successfully diagnose chronic obstructive pulmonary disease \cite{8682042}. Deep Learning has been used in recent work to predict heart failure risks \cite{info:doi/10.2196/20645}. The authors used a medical KG to guide the attentional mechanism of a Recurrent Neural Network trained with event sequences extracted from EMRs. Previously the authors of \cite{zhang2017hcnn} also predicted disease risk, but for a broader spectrum of NCDs, and using Convolutional Neural Networks in a KG of EMR events. Medical entity disambiguation is an NLP task aimed at normalizing KG entity nodes, and the authors of \cite{10.1145/3448016.3457328} approached this problem as one of classification using Graph Neural Network. Overall, multiple classical NLP methods have been applied to biomedical KGs, including biomedical KG forecasting from the point of view of link prediction (also known as literature-based discovery) \cite{crichton2019improving}.

\section{Conclusions}\label{sec:08_conclusions}

We trained different Transformer-based encoder-decoder NLMs to perform general-purpose reasoning as source task, including one-block and two-block NLMs. On the one hand, the performance results provided that one-block models showed the best metrics (accuracy and loss), however the divergence between train and test measurements was interpreted as overfitting at first glance. The two-block models, on the other hand, showed much more stable and less divergent (although decreasing) performance measurements, which was interpreted as higher generalization at first glance.

Nevertheless, hypothesis testing based on STS revealed with high confidence, on the one hand, that one-block models showed high generalization in source tasks, and from source to target tasks. These models managed to successfully distinguish themselves from simulated adversarial perturbation (the random baseline) without even learning directly from it ($p\ll 0.001$). On the other hand, two-block models ended by memorizing the source task, thus generating nearly random object phrases in the target task (only one two-block model barely reached $p<0.05$). This also confirmed that the NLMs' reasoning generalization may be manifested in semantic displacements and logical implications (as well as the abstraction and simplification of meanings), rather than in the ground truth distribution of output tokens.

We analyzed by inspection the semantic regularities shown by NLMs during source tasks and the NCD-related target task. The main behaviors observed as transferred were: $(i)$ the model generally takes concrete subjects and predicates, and associates with them the (semantically) most probable object, i.e. it learned semantic agreement; $(ii)$ the model ``objectifies'' and/or simplifies those meanings that could be considered subjective and/or complex (i.e., in terms of generalization, it is more likely to construct simpler than abstract and/or complex ideas). The two-block NLMs failed to acquire these skills, even the one that barely reached $p<0.05$ in the STS-based hypothesis testing.

Finally, our results suggest with significance that the predictions obtained by the proposed method can help answer research questions whose answers give rise to potential new knowledge, as we demonstrated for NCDs.

\section*{Acknowledgements}
We thank to PRODEP-SEP (project number UTMIX-PTC-069) for the support given. Also thanks to Dr. J. Anibal Arias Aguilar (Postgraduate Studies Division, UTM) for giving access to his computing nodes.

\bibliography{elsarticle-template}

\end{document}